\documentclass[journal]{IEEEtran}
\usepackage{graphicx}
\usepackage{caption}
\usepackage{subcaption}
\usepackage{xcolor}
\usepackage{url}
\usepackage{amssymb}
\usepackage{amsmath}
\usepackage{algorithm}
\usepackage{algorithmic}
\usepackage{float}

\usepackage{booktabs} 
\usepackage{rotating}
\usepackage{multirow} 
\usepackage{array} 
\usepackage{float}

\newcommand{\com}[1]{{\textcolor[rgb]{0.502, 0.502, 0.502}{#1}}}

\begin{document}
%
% paper title
% Titles are generally capitalized except for words such as a, an, and, as,
% at, but, by, for, in, nor, of, on, or, the, to and up, which are usually
% not capitalized unless they are the first or last word of the title.
% Linebreaks \\ can be used within to get better formatting as desired.
% Do not put math or special symbols in the title.
%\title{Learning Extended Body Schemas from Visual Keypoints for Object Manipulation}
% \title{Self-Supervised Learning of Extended Body Schemas for \textcolor[rgb]{0.0,0.6,0.5}{Multi-Modal?} Object Manipulation}
%\title{Multi-Modal Learning of Extended Body Schemas for Visual Object Manipulation}
%\title{Multi-Modal Keypoint and Model Learning for Visual Object Manipulation}
%\title{Multi-Modal Learning of Predictive Models for Visual Object Manipulation}
\title{Multi-Modal Learning of Keypoint Predictive Models for Visual Object Manipulation}
%\title{Multi-Modal Keypoint Learning for Visual Object Manipulation}
%\title{Multi-Modal Learning of Body Representations for Visual Object Manipulation}
%%%%%%%%%%%%%%%%%%%
% link for plots
% https://docs.google.com/presentation/d/1-7T3Nua3EL5HlKW3Bz6hdry2qG2BnG6lNYdSrgCGTYU/edit#slide=id.gd03456a37b_0_290
%%%%%%%%%%%%%%%%%%%
%
%
% author names and IEEE memberships
% note positions of commas and nonbreaking spaces ( ~ ) LaTeX will not break
% a structure at a ~ so this keeps an author's name from being broken across
% two lines.
% use \thanks{} to gain access to the first footnote area
% a separate \thanks must be used for each paragraph as LaTeX2e's \thanks
% was not built to handle multiple paragraphs
%

\author{Sarah Bechtle$^{1,2}$, Neha Das$^{2,3}$ and Franziska Meier$^{2}$
\thanks{$^{1}$Max Planck Institute for Intelligent Systems, Tübingen, Germany
        {\tt\small sbechtle@tuebingen.mpg.de}}%
\thanks{$^{2}$Facebook AI Research, Menlo Park, CA}%
\thanks{$^{3}$Technical University of Munich, Munich, Germany
        {\tt\small neha.das@tum.de}}%
}

% note the % following the last \IEEEmembership and also \thanks - 
% these prevent an unwanted space from occurring between the last author name
% and the end of the author line. i.e., if you had this:
% 
% \author{....lastname \thanks{...} \thanks{...} }
%                     ^------------^------------^----Do not want these spaces!
%
% a space would be appended to the last name and could cause every name on that
% line to be shifted left slightly. This is one of those "LaTeX things". For
% instance, "\textbf{A} \textbf{B}" will typeset as "A B" not "AB". To get
% "AB" then you have to do: "\textbf{A}\textbf{B}"
% \thanks is no different in this regard, so shield the last } of each \thanks
% that ends a line with a % and do not let a space in before the next \thanks.
% Spaces after \IEEEmembership other than the last one are OK (and needed) as
% you are supposed to have spaces between the names. For what it is worth,
% this is a minor point as most people would not even notice if the said evil
% space somehow managed to creep in.

% The paper headers
\markboth{Journal of \LaTeX\ Class Files,~Vol.~14, No.~8, August~2015}%
{Shell \MakeLowercase{\textit{et al.}}: Bare Demo of IEEEtran.cls for IEEE Journals}
% The only time the second header will appear is for the odd numbered pages
% after the title page when using the twoside option.
% 
% *** Note that you probably will NOT want to include the author's ***
% *** name in the headers of peer review papers.                   ***
% You can use \ifCLASSOPTIONpeerreview for conditional compilation here if
% you desire.

% If you want to put a publisher's ID mark on the page you can do it like
% this:
%\IEEEpubid{0000--0000/00\$00.00~\copyright~2015 IEEE}
% Remember, if you use this you must call \IEEEpubidadjcol in the second
% column for its text to clear the IEEEpubid mark.

% use for special paper notices
%\IEEEspecialpapernotice{(Invited Paper)}

% make the title area
\maketitle

% As a general rule, do not put math, special symbols or citations
% in the abstract or keywords.
\begin{abstract}
Humans have impressive generalization capabilities when it comes to manipulating objects and tools in completely novel environments. These capabilities are, at least partially, a result of humans having internal models of their bodies and any grasped object. How to learn such body schemas for robots remains an open problem. In this work, we develop an self-supervised approach that can extend a robot's kinematic model when grasping an object from visual latent representations. Our framework comprises two components: (1) we present a multi-modal keypoint detector: an autoencoder architecture trained by fusing proprioception and vision to predict visual key points on an object; (2) we show how we can use our learned keypoint detector to learn an extension of the kinematic chain by regressing virtual joints from the predicted visual keypoints. Our evaluation shows that our approach learns to consistently predict visual keypoints on objects in the manipulator's hand, and thus can easily facilitate learning  an extended kinematic chain to include the object grasped in various configurations, from a few seconds of visual data. Finally we show that this extended kinematic chain lends itself for object manipulation tasks such as placing a grasped object and present experiments in simulation and on hardware.
\end{abstract}

% Note that keywords are not normally used for peerreview papers.

% For peer review papers, you can put extra information on the cover
% page as needed:
% \ifCLASSOPTIONpeerreview
% \begin{center} \bfseries EDICS Category: 3-BBND \end{center}
% \fi
%
% For peerreview papers, this IEEEtran command inserts a page break and
% creates the second title. It will be ignored for other modes.
\IEEEpeerreviewmaketitle

\section{INTRODUCTION}
Humans have impressive generalization capabilities when it comes to manipulating objects and tools. These capabilities are, at least partially, a result of humans having internal models of their bodies \cite{hoffmann2010body} which enables them to predict the consequences of their actions. Endowing robots with similar capabilities remains an important open research problem.

\begin{figure}[t!]
    \centering
    \includegraphics[width=0.99\columnwidth]{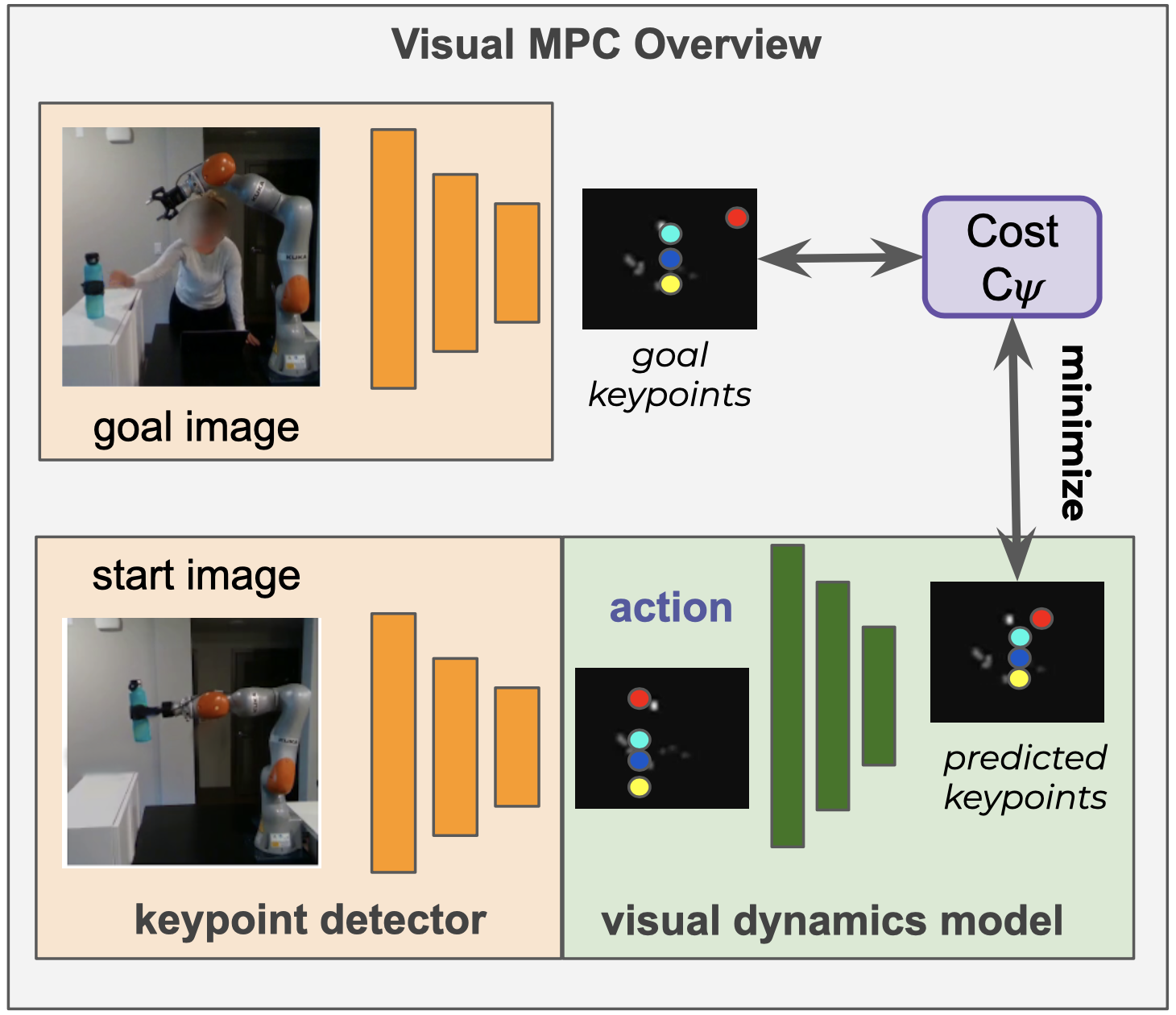}
   \caption{\small Overview of visual MPC via keypoints. Our work improves this framework in two places. (1) we present a multimodal version of the keypoint detector that merges vision and proprioceptive information, and (2) we show how we can use the visual keypoints to learn an extended kinematic chain that serves as reliable visual dynamics model for model based control.}
    \label{fig:visual_mpc_overview}
    \vspace{-0.2cm}
\end{figure}

In this work, we consider the problem of learning predictive models for visual control of grasped objects. Specifically, we consider the setting of visual model-predictive control for object manipulation as visualized in Figure ~\ref{fig:visual_mpc_overview}. In such settings, a low-dimensional representation of the object is extracted and then a predictive model is learned in that low-dimensional state-representation. This model is then used to optimize action sequences that accomplish a desired (visual) goal state.  We built on the recent success of learned keypoint representations which have been shown to capture task-independent visual landmarks for various applications \cite{minderer2019unsupervised, kulkarni2019unsupervised}. Benefits of such learned representations over more traditional object representations (such as $6D$ pose), are the ability to represent non-rigid objects, and not requiring object models. In contrast to other learned latent-state representations, keypoints are interpretable, and are less prone to become task-dependent. Given this learned state representation, current state-of-the-art then learns action-conditioned predictive models over keypoints \cite{minderer2019unsupervised, neha,manuelli2020keypointfuture}. Once such a dynamics model is trained, the robot can optimize actions to move observed keypoints into a desired goal keypoint configuration. 

While this overall framework is very promising, in the context of object manipulation at least two major challenges remain: 1) The self-supervised training of visual keypoints does not necessarily lead to keypoints that capture the object of interest. And even if it does, those keypoints may not consistently track the same part of the object throughout motion sequences;  2) In current state of the art methods \cite{minderer2019unsupervised, neha,manuelli2020keypointfuture}, keypoint predictive models are represented by unstructured neural networks, which tend to not extrapolate well to observations outside of the training distribution. As a result, such predictive models often do not perform well when used for object manipulation tasks.

%However, the question of how a robot can manipulate objects through keypoint representations remains an open challenging problem. 

We address these challenges as follows: First, during keypoint training, we encourage keypoint learning around the end-effector location, which leads to more consistent on-object keypoint detection. We achieve this by merging proprioceptive information, in form of joint positions, with visual information gathered from the camera. Then, instead of learning unstructured predictive models, we extend an existing kinematic model of the robots arm with virtual links and estimate the translation parameters of the links from visual keypoint predictions (see Figure~\ref{fig:extended_kinematic_chain}). This leaves us with an extended kinematic chain  or body schema that includes the object in  the  hand. Once the parameters of the virtual joints have been learned, we have a fully differentiable forward kinematics model that can be utilized for model-based control. A key-feature of this approach is that we can estimate parameters to adapt to various grasp variations of the object. We will show experimental evaluation of these benefits in section \ref{sec:experiments}.

To summarize, in this work we propose a fully self-supervised framework to automatically learn extended body schemas from multi-modal data to successfully perform object manipulation tasks. Towards this our contributions are as follows:
\begin{enumerate}
    \item We propose a multi-modal keypoint learning approach that merges proprioceptive state, RGB and depth measurements to spatially bias the visual keypoints towards the end-effector. This leads to better on-object keypoint predictions as compared to keypoint methods \cite{minderer2019unsupervised} that only take RGB measurements into account.
    \item We use a fully differentiable kinematic representation of the manipulator that we extended with a priori unknown virtual links and joints representing the object. To recover the virtual joints parameters, we propose a gradient based learning approach that learns the parameters given the visual keypoint predictions as targets.
    \item We evaluate the learned extended kinematic chain on a downstream object manipulation task and show that our self-supervised approach achieves good performance in simulation and on hardware.
    \item We evaluate our multimodal keypoint detector and extended kinematic chain on a 7DoF iiwa Kuka arm that has various objects in hand, in simulation and on hardware.
\end{enumerate}
 Our results show that when trained with proprioceptive information, the learned keypoints represent the manipulated objects more reliably. After regressing the virtual joints from the visual keypoint information, we compare the resulting extended kinematic model to dynamics models learned with neural networks, on a model-based control task: placing a grasped object. Our method outperforms the learned dynamics models by an order of magnitude on the downstream task.
\section{Related Work}\label{sec:related_work}
\subsection{Internal Representations of Body in the  Brain}
It has been well established that distinct representations of the body are created, adapted and utilized by the brain while performing sensorimotor tasks and throughout one's lifetime (See \cite{hoffmann2010body} for a review). Research in the domain of cognition \cite{limanowski2020active,van1999integration,sober2005flexible} has additionally inferred that the "internal state" of a limb is informed by both vision and proprioception, and that both signals are combined in a weighted fashion to update this internal representation, with the weights attributed to each signal depending on their reliability. 
Several works (\cite{cardinali2009tool,cardinali2012grab,baccarini2014tool,martel2021long}) also consider the plasticity or extension of body schemas during (or in anticipation of) tool use in human and primate subjects. \cite{martel2021long} shows for instance that the use of a 40 cm long tool in a study causes the participants to move as though their arm is longer, indicating a change in their body schema.
Taking inspiration from biological cognition, we combine the above notions in our approach. We extend the existing kinematic chain (analogous to body schema in humans) of our robotic arm to include a grasped object. Furthermore, we fuse signals from proprioception and vision for estimating the state of the extended parts of the chain more accurately. We expand more on related works in robotics in these contexts in sections \ref{sec_rel:vision_proprioception} and \ref{sec_rel:body_schemas}.
\begin{figure*}[ht]
    \centering
    \includegraphics[width=0.99\textwidth]{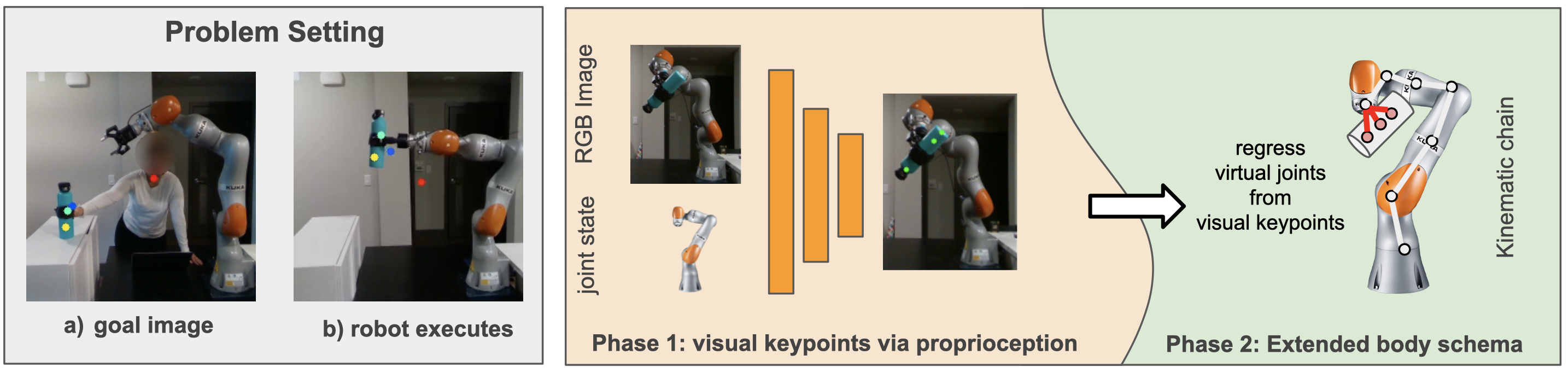}
   \caption{\small Overview of self-supervised learning of extended body schemas. Our approach comprises two phases, (1) in the first phase we learn an autoencoder architecture to detect visual keypoints on the object in the manipulator's hand by merging proprioceptive and visual information. The forward kinematic model of the robot is used, to create a kinematic features map. The visual and the kinematic feature maps are then combined to train the keypoint detector. In the (2) second phase we use the predicted visual keypoints, that ideally are detected on the object, to learn an extended kinematic chain of the manipulator that inherently includes the object in the robot's hand. The extended kinematic chain is then used to accomplish a manipulation task. Learning happens fully from visual and proprioceptive information and can adapt to different rigid bodies and grasps.}
    \label{fig:extended_kinematic_chain}
    \vspace{-0.5cm}
\end{figure*}
\subsection{Learning Visual Representations for Model Predictive Control (MPC)} \label{sec_rel:rep_learning}
Since the core emphasis of the article is learning representations for robotic object manipulation, we contrast our framework to other relevant approaches with respect to latent space representation, dynamics model learning and action optimization. Approaches that employ model-predictive control in visual space can be distinguished by how much, and what kind of structure is infused into the predictive model. Approaches such as \cite{ebert2018_visual_foresight}, utilize no structure and learn a predictive model $f$ directly in pixel space.
Both, \cite{watter2015embed} and \cite{byravan_2018_se3net_control} learn a latent representations $z$ and a predictive model $f$ in the latent-space. While \cite{watter2015embed} assumes no structure for learning $z$, they assume locally linear dynamics for learning the latent space transition model $f$. \cite{byravan_2018_se3net_control} on the other hands learns a structured SE3 latent representation that is more interpretable, and learns unstructured predictive models $f$.  
Our method builds upon an self-supervised keypoint representation learning approach in \cite{minderer2019unsupervised, kulkarni2019unsupervised}, which \cite{neha, DIGIT} utilize for object manipulation tasks. It relies on an auto-encoding scheme with a structural bottleneck that can extract 2D keypoints representing the object(s) of interest. However, as observed by \cite{neha, DIGIT}. the keypoints encoded by this method are not always consistently on the object of interest. We illustrate this further in our experiments that compare this keypoint detector \cite{minderer2019unsupervised} with our multimodal keypoint detector.

\cite{manuelli2020keypointfuture} is another recent work that takes a structured approach towards learning keypoints and utilizes them for model-predictive control. The keypoints learned are, by design, always located on the object and are consistent across frames. This is achieved by first learning a dense visual object descriptor \cite{8917583} for the object of interest.  
which has two requirements: (a) pixel to pixel correspondences between image sets, and (b) an object mask learned by utilizing results from \cite{finman2013toward}. Following the training of a dense visual descriptor, $k$ descriptors are sampled as keypoints. An unstructured predictive model $f$ is then learned in keypoint space \cite{manuelli2020keypointfuture}. In contrast to this approach for keypoint detection, our multimodal detector does not need pixel-wise corresponding images for supervision or an object mask, while still being able to produce keypoints that are reliably present on an object (\ref{sec_rel:vision_proprioception}).

The above mentioned approaches, also differ in how action sequences $u$ are optimized: via the cross-entropy method CEM \cite{CEM, ebert2018_visual_foresight, DIGIT, manuelli2020keypointfuture}, gradient based optimization \cite{byravan_2018_se3net_control, neha}, Model Predictive Path Integrals \cite{manuelli2020keypointfuture} or by using optimal control methods \cite{watter2015embed}. Our framework is agnostic to the specific action optimization method. We present results using gradient based action optimization.
\subsection{Multi-Modal Learning: Fusing Vision and Proprioception}\label{sec_rel:vision_proprioception}
In the previous subsection we discussed model-predictive control that utilize visual data only. However, there is more and more evidence that utilizing multi-modal sensor streams improves perception and manipulation \cite{bohg2017interactive,edelman1987neural,lacey2016crossmodal}. This has been explored in various robotics applications. For instance, fusing vision and proprioception \cite{cifuentes_articulated_tracking_2017,kappler2018real,martin2017cross} or combining visual and tactile information \cite{martin2017cross,lambert2019joint,yu2018realtime} has been explored for better state-estimation in applications such as object-tracking. These approaches are mostly concerned with understanding state-estimation from multi-modal sensor data and assume expert-designed low-dimensional features are extracted from each modality. Manually designing features for heterogeneous data is extremely challenging, time-consuming and thus not scalable. Because of this reason, there has been a recent push towards learning state representations from multi-modal data for a wide range of applications. For example, there have been many works that have explored the correlation between auditory and visual data for tasks such as speech or material recognition or for sound source localization \cite{ngiam2011multimodal,owens2018audio,owens2016visually,yang2017deep}. Several work \cite{bekiroglu2011learning,calandra2018more,gao2016deep,sinapov2014learning} fuse visual and haptic data for various applications such as grasp stability assessment, manipulation, material recognition, or object categorization. 

As discussed above, fusing multiple sensor modalities for better state-estimation is common in the rigid-body tracking literature, and recently it has also been shown by \cite{lee2019making} that it leads to better learned state representations for robotic manipulation tasks. In this work we show how fusing proprioception and vision (rgb images and depth) significantly improves the learning of keypoint representations.

\begin{figure*}[ht]
\centering
\includegraphics[width=0.9\textwidth]{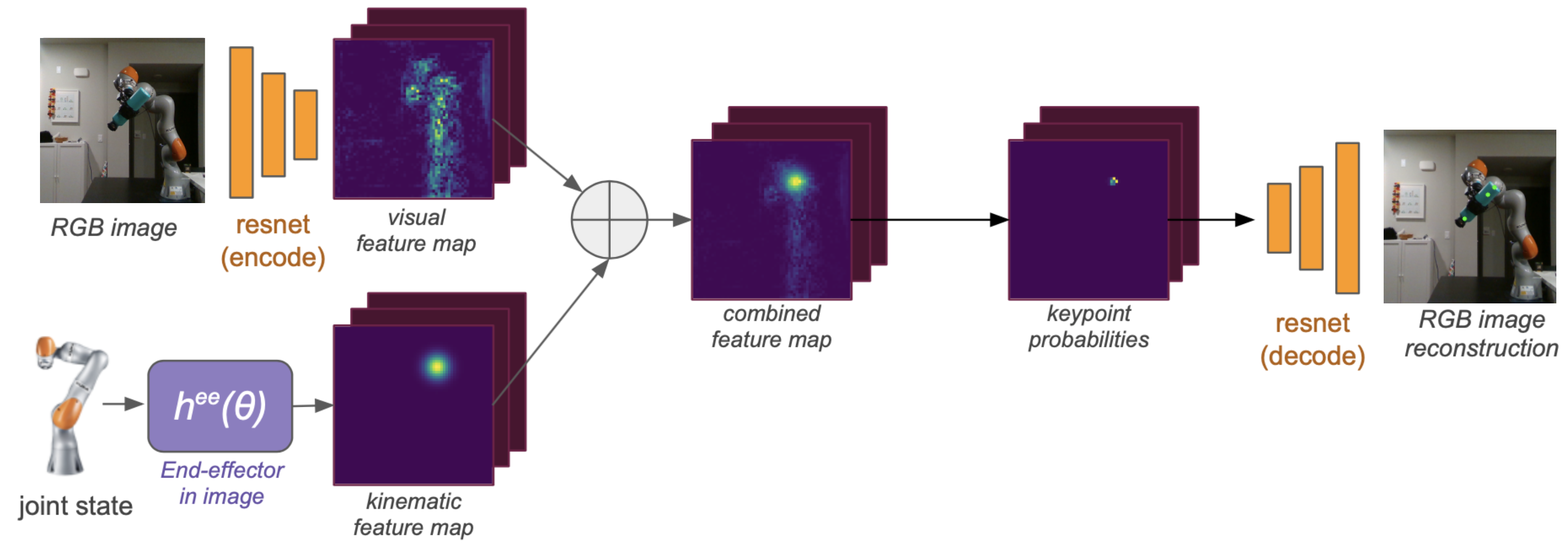}
\caption{\small Phase 1: Multimodal keypoint detector, fusing vision and prorioception to predict visual keypoints. The forward kinematic model of the robot is used, to create a kinematic features map. The visual and the kinematic feature maps are then combined to train the keypoint detector. 
}
\vspace{-10pt}
\label{fig:unstr_keypoint_det}
\end{figure*}

\subsection{Learning body schemas} \label{sec_rel:body_schemas}
A promising alternative to learning unstructured predictive models are approaches that use structure for body schema learning. A body schema \cite{hoffmann2010body} is a representation of a robot's body and its extensions such as grasped objects or tools, which can then be used for control. Body schema approaches can be classified into approaches that estimate parameters of structured representations of a body and tool (i.e kinematic representation) \cite{sturm2009body, martinez2010body, ulbrich2009rapid, hersch2008online, gothoskar2020learning, stepanova2019robot}, or of unstructured models, such as neural network representations \cite{boots2014learning, hikita2008visual, rolf2010learning, nabeshima2006adaptive, yoshikawa2003does, schillaci2012coupled}.
To address the challenge of learning models that generalize to differet parts of the state space, we follow approaches that learn parametrized kinematic models. Prior work in this category typically utilizes markers or ground truth knowledge about the end-effector or tool-tip location in the robots workspace \cite{sturm2009body, martinez2010body, ulbrich2009rapid, gothoskar2020learning} or simplified visual signals \cite{hersch2008online}, and instead focus on learning kinematic parameters. In contrast to this, our work extends an existing kinematic chain to include a grasped object purely from learned visual latent representations in a self supervised way. 

\section{Problem Setting and Method Overview}

In this work, we address deterministic, fixed-horizon and discrete-time control problems with continuous states $\mathbf{s} = (s_1, \dots, s_T)$ and continuous actions $\mathbf{u} = (u_1, . . . , u_T)$. Each state
$s_t = [\theta_t, z_t]$ is the concatenation of the measured joint angles $\theta_t$ and a learned visual latent state $z_t$ at time step $t$. To solve the control problem we use a learned visual predictive dynamics model $\hat{s}_{t+1} = f(s_t, u_t)$ and a cost function $C(z_t, z_\text{goal})$ that measures the distance between current and desired goal state in the visual latent space. 

The learned predictive model $f$ predicts the change in object state, as perceived from the camera, given the current joint displacements. $f$ is learned in a self supervised fashion by merging proprioceptive and visual information and we do not assume any additional model knowledge of the external object. 

Learning $f$ involves two phases. In the first phase a keypoint detector is trained to detect keypoints $z$ on the object in the manipulator's hand from the visual data. The keypoint detector follows an autoencoder architecture and we present a novel approach to merge visual and proprioceptive information for keypoint detector training, when encoding the visual information. In the second phase of our approach, we use the learned keypoint detector to learn the parameters of an extended kinematic chain of the robot. The extended kinematic chain, extends the standard kinematics of the robot arm to include also the object in the manipulator's hand. The extended kinematic chain can then be used to optimize a control policy for a manipulation task. The control tasks are characterized by a desired goal state for an object in the manipulators hand. The goal state is provided in the visual latent space i.e. the visual keypoits. An overview of our self-supervised learning approach can be seen in Fig. \ref{fig:extended_kinematic_chain}.

\section{Phase 1: Fusing Proprioception and Vision for Multi-Modal Keypoint Learning}\label{sec:approach_keypoint}
In this section we introduce a novel multimodal keypoint learning framework that leverages RGB, depth and proprioceptive measurements for improved keypoint training and prediction. Intuitively, proprioception is the sense of self movement and body position. In the context of this work we use robot joint positions and forward kinematics as proprioceptive information during learning. 

\subsection{Learning Visual Keypoints for Object Manipulation}
We base our multi-modal keypoint learning framework on the visual keypoint detector presented in \cite{minderer2019unsupervised}. To learn keypoints \cite{minderer2019unsupervised} uses an autoencoder with a structural bottleneck to detect 2D visual keypoints that correspond to pixel positions  with maximum variability in the input data. For keypoint prediction \cite{minderer2019unsupervised} extract $K$ 2-D visual feature maps $o_k^{\text{visual}}$ through a mini-RESNET 18, where $K$ is the number of keypoints.
The autoencoder architecture is trained with RGB image sequences collected from videos. The maximum variability can intuitively be thought of as areas of biggest movement in the image. The autoencoder architecture is then trained with a combination of losses. The loss is composed of a reconstruction error ($\mathcal{L}_\text{rec}$), a keypoint sparsity error ($\mathcal{L}_\text{spa}$), and a keypoint separation loss ($\mathcal{L}_\text{sep}$), that encourages the separation of keypoints in pixel-space.
\begin{equation}
  \mathcal{L}_\text{\cite{minderer2019unsupervised}} = \mathcal{L}_\text{rec} + \lambda_\text{spa} \cdot \mathcal{L}_\text{spa} + \lambda_\text{sep} \cdot \mathcal{L}_\text{sep}   
\end{equation}
where $\lambda_\text{spa}$ and $\lambda_\text{sep}$ represent the scale parameters for various losses.

In the next sections we are going to present how we extend the keypoint detector by \cite{minderer2019unsupervised} to a multimodal keypoint detector that merges vision and proprioception.
\subsection{Fusing visual and kinematic feature maps}\label{sec:forward_kin}
We now present how we include proprioceptive information when training our multimodal keypoint detector. Our method extends the one presented by \cite{minderer2019unsupervised} in two places. We (1) create, additionally to the visual feature map, a kinematic feature map, representing the end effector position of the robot in image space and (2) we extend the training loss with a kinematic consistency loss. In the following we are going to explain both extensions in detail.

To create the kinematic feature map we specify $s_n^\text{img} = h^n(\theta)$ as a forward kinematic function, that directly projects the 3D position of link $n$ into image space when the robot is in configuration $\theta$. 
To retrieve $s_n^\text{img}$, first, the robot's forward kinematic model is used to deliver $x_{n}$, the 3D location of link $n$ in the robot's coordinate frame. Then $x_n$ is projected into image and depth space. The full transformation is given by
\begin{equation*}
     s_n^\text{img,depth} = h^n(\theta) = T_\text{proj}T_\text{cam} x_{n} \text{ where } x_n = \prod_{i=1}^n T_i(\theta_i; \phi_i)
\end{equation*}
where $T_i$ is the transformation matrix from the coordinate frame of link $i$ to link $i-1$ and $\phi_i$ are the parameters (rotation and translation) of link $i$;  $T_\text{cam}$ transforms $x_n$ from robot to camera coordinate frame, and $T_\text{proj}$ performs the projection to image and depth space. In the following, we will use $s_n^\text{img, depth}$ to denote the use of both image and depth value, and $s_n^\text{img}$ when we only use image pixel predictions. In the following, we assume that the parameters $\phi_i$ up to the end-effector index $n=\text{ee}$ are known, and we will use $h^\text{ee}(\theta)$ to combine visual and proprioceptive information to train the keypoint detector.

In order to spatially bias keypoint learning to be close to end-effector location, we propose to include proprioceptive information during keypoint detector training in the encoding phase. We do this by computing a second feature map, which we call the kinematic feature-map $o^\text{kin}$, visible in Fig.\ref{fig:unstr_keypoint_det}. The kinematic feature-map is generated by first computing the end-effector position in image space, $s_\text{ee}^\text{img} = h^\text{ee}(\theta)$, as described in section \ref{sec:forward_kin}, followed by placing a Gaussian blob $g^\text{kin}$ over the location of the projected end-effector position in image space:
\begin{equation*}
    o^\text{kin} = g^\text{kin}(s_\text{ee}^\text{img}) %= \mathcal{N}(s^\text{img}_{\text{ee}},\Sigma)
\end{equation*}
where $g^\text{kin}$ converts the pixel coordinates of $s_\text{ee}^\text{img}$ into a heatmap with a Gaussian shaped blob $\mathcal{N}(s^\text{img}_{\text{ee}},\Sigma)$ centered at $x,y$ pixel locations of $s_\text{ee}^\text{img}$. 

We then combine the kinematic with the visual feature map, creating a joint feature-map  $o^\text{joint} = o^\text{visual} + o^\text{kin}$, which fuses visual and proprioceptive information. Given $o^\text{joint}$, keypoints are trained through a reconstruction objective, similar to the method presented in \cite{minderer2019unsupervised}. Figure~\ref{fig:unstr_keypoint_det} provides an overview of our adapted encoder-decoder architecture. 

\subsection{Utilizing proprioception and depth to augment loss}\label{sec:structured_keypoint_training}
We extend the loss proposed in \cite{minderer2019unsupervised} $\mathcal{L}_\text{\cite{minderer2019unsupervised}}$ by including a term that penalizes the distance between keypoints and end-effector. This \emph{kinematic consistency loss}, together with the kinematic information during feature map creation, spatially biases the keypoint learning towards the end-effector location to incentivize the detector to place keypoints on the object.
The kinematic consistency loss, is given by
\begin{equation}
    \mathcal{L}_{\text{kin}} = \sum_{k} (z_k^{[x,y,depth]}-s^\text{img,depth}_{\text{ee}})^2
\end{equation}
where $z_k^{[x,y,depth]}$ is the $x,y$ pixel locations and depth value of the predicted keypoint $z_k$. The depth value of $z_k$ is retrieved, by querying the depth image at pixel locations $x,y$. 

The complete loss that is minimized during Phase I is thus given by:
\begin{equation}
    \mathcal{L} = \mathcal{L}_\text{\cite{minderer2019unsupervised}} + \lambda_\text{kin} \cdot \mathcal{L}_{\text{kin}}
\end{equation}
where $\lambda_\text{kin}$ again is a scaling parameter for kinematic consistency loss.

To train our keypoint detector we collect visual and proprioceptive data $\mathcal{D}_\text{key-train}$ for self-supervised keypoint training. 
After this training phase, we have a keypoint detector that predicts keypoints $z$ of dimensionality $K \times 3$. Here $K$ is the number of keypoints, and each keypoint is given by $z_{k} = ({z}_{k}^{x}, {z}_{k}^{y}, {z}_{k}^{\text{depth}})$, where ${z}_{k}^{x}, {z}_{k}^{y}$ are pixel locations of the $k-\text{th}$ keypoint, and ${z}_{k}^{\text{depth}}$ is the value of the depth image at location ${z}_{k}^{x}, {z}_{k}^{y}$. 

\subsection{Inference at test time}
It is important to note that the proprioceptive information is only added during keypoint detector training. Also during training, we do not need to necessarily include proprioceptive information for each datapoint, but can also train from mixed datasets that contain only visual information and visual and proprioceptive information. After training, the proprioceptive information is no longer needed for keypoint prediction at test time since the proprioceptive information is now inherently part of the leared keypoint detector. The keypoint detector can be used with purely visual input. This allows extraction of keypoints from a goal image that is produced, for instance, by recording human demonstrations. At inference time, the keypoint detector predicts the $x,y$ locations of the keypoint $z_k$, the depth value is then retrieved by querying the depth image at pixel location $x,y$.

\section{Phase 2: Learning Body Schema Extension from Visual Keypoints}\label{sec:approach}
The multimodal keypoint detector presented in section \ref{sec:approach_keypoint} merges visual and proprioceptive information during training, facilitating better on-object keypoint predictions during test time. Since the keypoints predicted by the detector are more likely placed on the object in the manipulator's hand, they will enable us to learn an extended kinematic chain of the robot arm that includes the object in the manipulator's hand. The extended kinematic chain can then be used for model-based control on a downstream manipulation task. In this section we are going to explain how we learn the extended kinematic chain and how it can be used for a downstream manipulation task.

\begin{figure}
    \centering
    \includegraphics[width=0.49\textwidth]{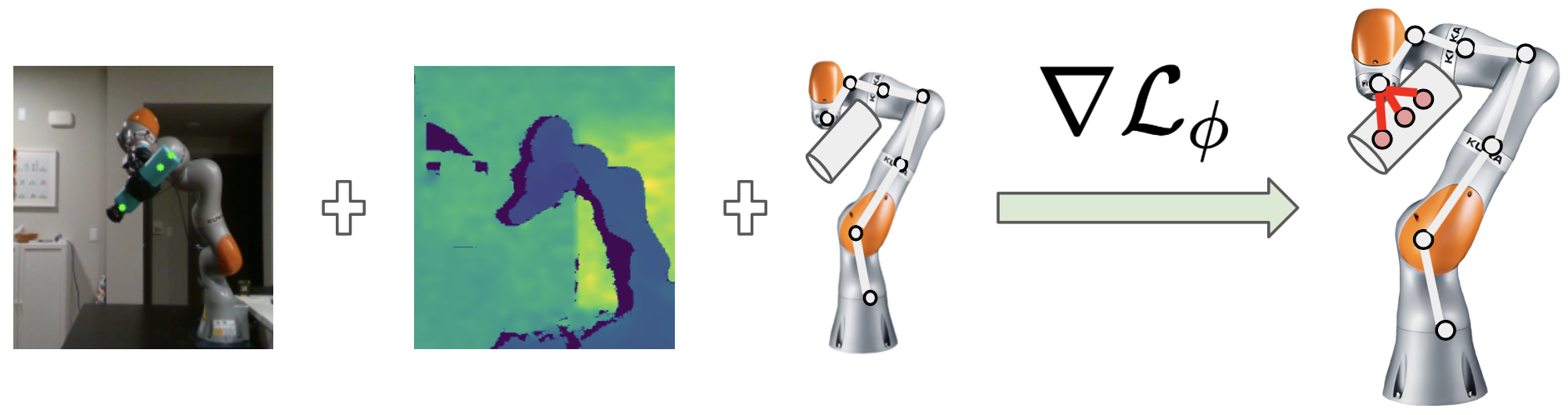}
    \caption{\small Phase 2: The translation parameters of the virtual joints representing the object in the manipulator's hand (red dots) are estimated from the x,y pixel location of the visual keypoints together with the depth information of the pixels. The virtual joints extend the kinematic chain of the robot to also include the object, as is illustrated in the figure.}
    \label{fig:phase_2_overview}
\end{figure}

As already pointed out in section \ref{sec:related_work}, an object or a tool in the gripper of a manipulator can be seen as an extension of the kinematic chain of the robot. In Fig. \ref{fig:phase_2_overview}, we visually show how the extension of the kinematic chain looks like for the kuka manipulator: additional virtual links and joints are added, representing the object in the manipulator's hand, to create a full chain that includes the object. The extended kinematics of the robot can then be controlled. 

In order to learn this extended body schema from vision we use the keypoint predictions of the multimodal keypoint detector presented in section \ref{sec:approach_keypoint}. The predicted keypoints on the object are then used to regress the kinematic parameters of the extended body schema.

\subsection{Estimating virtual joints from visual keypoints}\label{sec:estimating_joints}
During the second phase of our approach, the extended kinematic chain of the manipulator should be learned to include the object in the manipulator's hand. The self-supervised nature of our approach allows the kinematic chain to adapt when the object changes or the grasp around the object shifts, making the learning of the extended kinematic chain flexible to changes in the experimental setup.
Once the keypoint detector has been trained, we use its predictions to learn an extension of the kinematic chain to include the object into the body schema. We extend the kinematics model from Section~\ref{sec:forward_kin}, to include virtual joints, one per visual keypoint, such that  $s^\text{img,depth}_\text{k} = h^k_\phi(\theta)$ is the projection of the virtual joint in image space. Here, $\phi$ are the learnable translation parameters of the virtual joints, that we aim to learn such that they represent the object (see Fig. \ref{fig:extended_kinematic_chain} and Fig. \ref{fig:phase_2_overview}, where the red circles represent the virtual joints with the virtual links connecting them to the robot end effector, we assume the extension of the kinematic chain starts at the end effector). To regress $\phi$, a dataset $\mathcal{D} = \{ (x_t=\theta_t, y_t=z_t)\}_{t=1}^T$ is collected where the trained keypoint detector is used to predict visual keypoints $\mathbf{z}^{[x,y,depth]}$ on images of the object in the robot's hand while being in joint configuration $\theta$. The goal of this learning problem is to regress the translations $\phi$, such that the output of $h^k_\phi(\theta)$ matches the  keypoints detected by the detector. In other words, to optimize $\phi$ we want to minimize the loss $\mathcal{L}_{\text{trans}}$ via gradient descent. 
\begin{equation}
    \mathcal{L}_{\text{trans}} = (\mathbf{s^\text{img,depth}_{\text{k}}} - \mathbf{z}^{[x,y,depth]}_{\text{k}})^2
\end{equation}
where $\mathbf{s^\text{img,depth}_\text{k}} = h^k_\phi(\theta)$. This loss is a function of the learnable kinematic parameters $\phi$, making it possible to update $\phi$ via gradient descent.
Learning the translation parameters of the virtual joints, results in a new extended kinematic model, which includes the object. This new kinematic chain does not require visual information anymore and can be used for action optimization. 
As will be presented in section \ref{sec:experiments} using our version of the keypoint detector, that merges visual and proprioceptive information provides more accurate on-object keypoint predictions and thus makes learning the extended kinematic chain possible.

\section{Gradient-Based Control for Object Manipulation}\label{sec:gradient_based_control}
The goal is to successfully perform a manipulation task in the visual domain. This means that the desired goal position of the object in the manipulator's hand is given in image space instead of in joint space. Specifying a desired goal position in image space is more intuitive compared to specifying it in joint positions, where the relationship between joint position and real world object position is not readily available. 
In contrast to other visual MPC work \cite{ebert2018_visual_foresight, byravan_2018_se3net_control,neha}, that utilize learned visual dynamics models to optimize actions $\mathbf{u}$, we make use of our learned extended kinematic chain. Specifically, we define a visual predictive model $s_{t+1} = f(s_t,u_t)$ where $s_{t+1} = [\theta_{t+1},z_{t+1}]$, and
\begin{align}
    \theta_{t+1} &= \theta_t + u_t\\
    z_{t+1} &= h^k_{\phi}(\theta_{t+1})
\end{align}
where actions $\mathbf{u_t}$ are desired changes in joint positions and $h$ is, as defined in section \ref{sec:approach_keypoint}, the forward kinematic call of the learned extended kinematic model projected in image space.
To optimize actions, we follow the gradient based action optimization presented in \cite{byravan_2018_se3net_control, neha} and minimize a task specific cost function $C$. Specifically, to optimize a sequence of action parameters $\mathbf{u} = (u_0, u_1, \dots, u_T)$ for a horizon of $T$ time steps, we first predict the trajectory $\hat \tau$, that is created using an initial  $\mathbf{u}$ from starting configuration $s_0$: 
$\hat{s}_{1}= f(s_0,u_0), 
    ~\hat{s}_{2} =  f(s_1,u_1), 
    \dots
    ~\hat{s}_{T} = f(s_{t-1},u_{t-1}),
$
which generates a predicted (or planned) trajectory $\hat \tau$. 

Practically, this step uses the extended kinematics model $h^n_\phi(\theta)$ to simulate forward what would happen if we applied action sequence $\mathbf{u}$ to the initial state $s_0$. We then measure the cost achieved $C(\hat \tau, z_\text{goal})$, where $z_\text{goal}$ is a goal location in the visual domain. Since $h$ is differentiable, the cost of the planned trajectory can be minimized via gradient descent by taking the gradient of the cost with respect to the action sequence and performing a gradient update step.
\begin{align}
    \mathbf{u}_\text{new} = \mathbf{u} - \eta \nabla_u C(\hat \tau, z_\text{goal})
\end{align}
Algorithm \ref{algo:gradient_based_control} shows the details of our visual MPC algorithm.
\begin{algorithm}[H]
% \vspace{-1cm}
\begin{algorithmic}[1]
\footnotesize{
\STATE{$f(s_t,u_t)=[\theta_t+u_t,h^k_{\phi}(\theta_{t+1})]$}
\STATE{initial state $s_0 = [\theta_0, z_0]$}
\FOR{each $epoch$}
\STATE{$u_t = 0, \forall t=1,\dots, T$}
\STATE{\com{// rollout $\hat{\tau}$ from initial state $s_0$ and actions $u$}}
\STATE{$\hat{\tau} \gets \text{rollout}(s_0, u, f)$}
\STATE{\com{// Gradient descent on $u$}}
\STATE{${u}_{new} \gets {u} - \eta \nabla_{{u}}C(\hat{\tau}, z_{goal})$}
\ENDFOR
}
\end{algorithmic}
\caption{\strut\small Gradient Based Control}
\label{algo:gradient_based_control}
\end{algorithm}

\section{Experiments: Self-Supervised Learning of Body Schemas}\label{sec:experiments}
In this section, we present the experimental evaluation of our approach. We train two versions of the keypoint detector: our multimodal keypoint detector, merging vision and proprioception, following the approach presented in section \ref{sec:approach} and the detector presented in \cite{manuelli2019kpam},\cite{minderer2019unsupervised}, as a baseline comparison. All of our experiments are performed on a 7 DoF iiwa Kuka arm \cite{kuka}, with an object attached to the Kuka's gripper. We present experiments in simulation and on hardware. For our simulation experiments, we use the Habitat simulator \cite{habitat} with a pybullet integration \cite{pybullet}. 
In simulation, we use three different objects to show the adaptability of our approach. The experimental setup together with the objects used, is shown in Fig. \ref{fig:experimental_setup_objects} for simulation and hardware.
\subsection{Data Collection}
\begin{figure}[h]
    \centering
    \includegraphics[trim=50 100 150 100,clip,width=0.24\columnwidth]{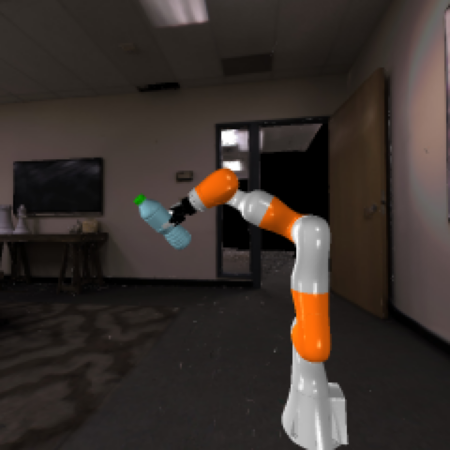}
    \includegraphics[trim=50 100 150 100,clip,width=0.24\columnwidth]{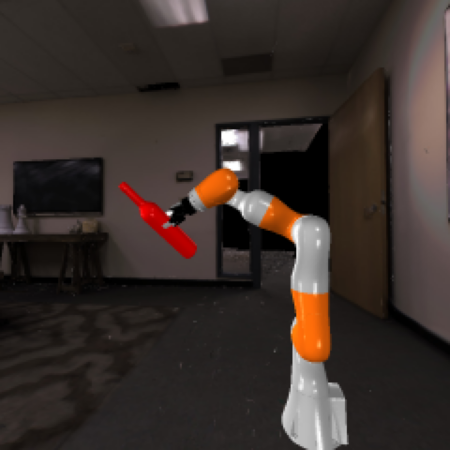}
    \includegraphics[trim=50 100 150 100,clip,width=0.24\columnwidth]{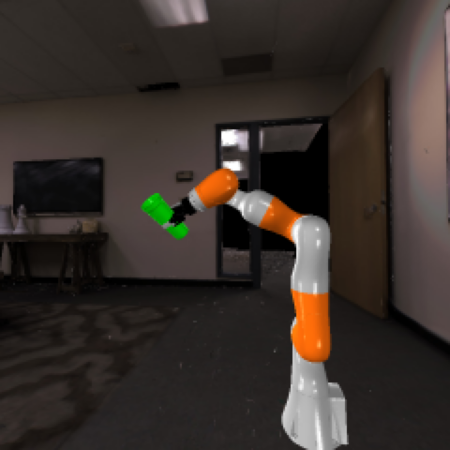}
    \includegraphics[trim=50 100 150 100,clip,width=0.24\columnwidth]{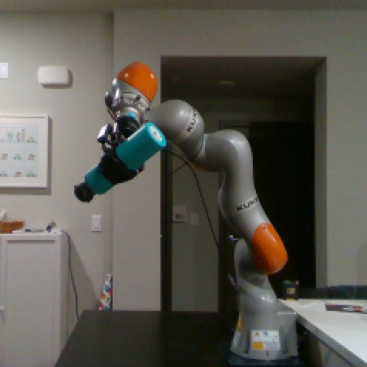}
    \caption{The 3 objects used for simulation experiments, and one object on hardware.}
    \label{fig:experimental_setup_objects}
    \vspace{-0.5cm}
\end{figure}
The keypoint architecture as presented in \cite{minderer2019unsupervised} is built to learn keypoints that capture motion in video-sequences. Because we aim to learn keypoints on the object, we collect data that only has object motion, similar to \cite{neha}. Specifically, we first move the manipulator to a random joint configuration, then we keep the manipulator in its position and only move the end-effector for 6 seconds.  In total we move the robot to 60 initial random joint configurations, and from there collect data at a frequency of 5Hz for a total of 5 seconds, for each object. We collect image data, which includes RGB and depth data, alongside with proprioceptive information about the joint positions.
Data collection on hardware follows the same procedure as presented in \cite{neha}: we collect $50$ sequences of motion data in which only the end-effector and object move, starting from a random joint configuration. Each sequence is $3$ seconds long, and contains $10$ data frames. 
In simulation we perform this data collection for 3 objects and for 1 object on hardware, as shown in  Figure~\ref{fig:experimental_setup_objects}.

\begin{figure*}[ht!]
    \centering
    \includegraphics[width=0.2\textwidth]{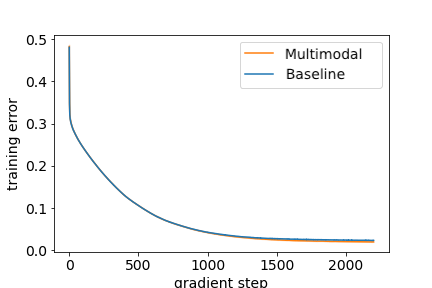}
    \includegraphics[width=0.12\textwidth]{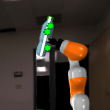}
    \includegraphics[width=0.2\textwidth]{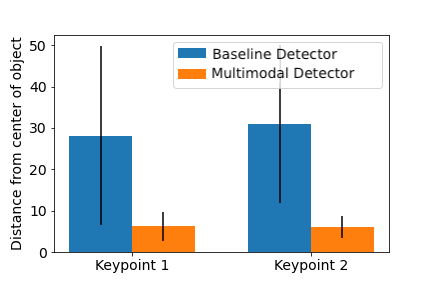}
    \includegraphics[width=0.2\textwidth]{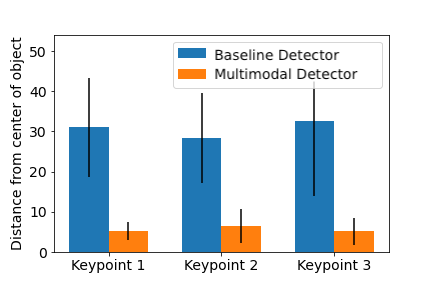}
    \includegraphics[width=0.2\textwidth]{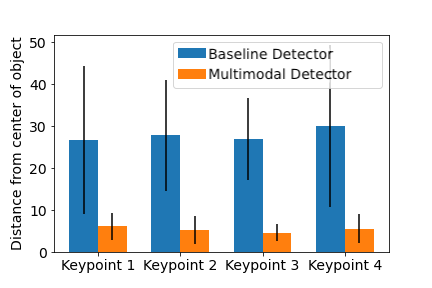}
            \includegraphics[width=0.2\textwidth]{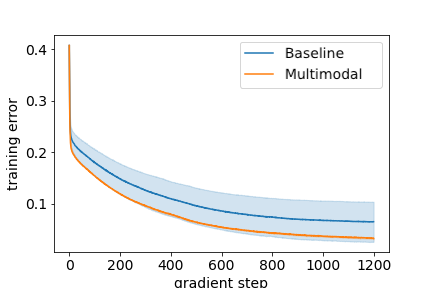}
        \includegraphics[width=0.12\textwidth]{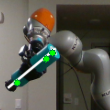}
    \includegraphics[width=0.2\textwidth]{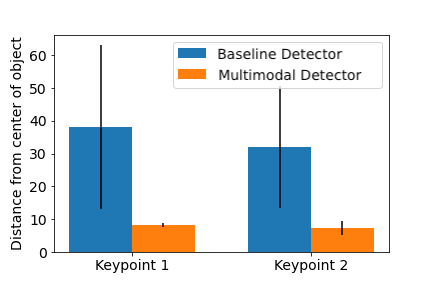}
    \includegraphics[width=0.2\textwidth]{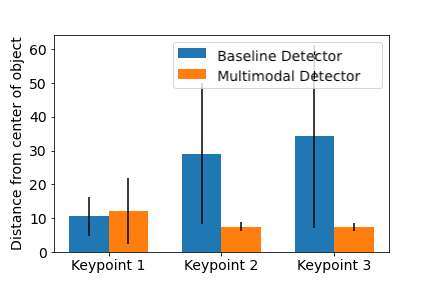}
    \includegraphics[width=0.2\textwidth]{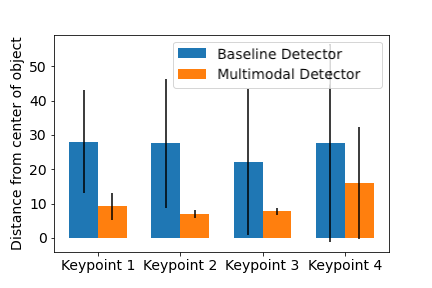}
    \caption{\small {\bf (top)} simulation results; {\bf (bottom)} hardware results. {\bf (1st col)} training loss for detector training, both detectors converge during training. {\bf (2nd col)} illustration of evaluation metric. {\bf (3rd to 5th col)} Results for training keypoint detectors with $K=2,3,4$ keypoints, respectively. The bar plots show the average distance from the center line for each keypoint in pixel space. The multimodal keypoint detector (orange) predicts keypoints that are closer to the center line, and the average pixel distance suggests that they are placed on the object.}
    \label{fig:comp_detectors_training}
\end{figure*}

\subsection{Does proprioception help train better keypoints?}
In this section we want to analyse how merging proprioceptive information and visual information benefits on-object keypoint detection. For this purpose we train two kinds of keypoint detectors, one that includes proprioceptive and visual information which we call the multimodal keypoint detector and one that only uses visual information which is our baseline and equivalent to the keypoint detector presented in \cite{minderer2019unsupervised}.
After training, we compare the performance of the multimodal and baseline keypoint detectors of achieving on-object keypoint detection. For our simulation experiments, we present results averaged over five seeds and 3 different objects, where we train a detector per object. On hardware, the experiments are also averaged over five seeds and we present results for two to four keypoints on a single object. 
First, we show the training loss curves in Figure~\ref{fig:comp_detectors_training}(left), averaged across seeds, objects and number of keypoints. In simulation, the loss curves are very similar, we believe that this is due to the reconstruction loss being the biggest component of the total training loss. Since only the gripper and object moves, the reconstruction for most of the image works well, reducing the loss and resulting in low training error even if the keypoints are not placed on the object.

Next, we evaluate how well the learned keypoints capture moving objects. To this end, we define an imaginary line, that cuts through the object in the middle and connects its extremities. A visualization of it is visible in Fig. \ref{fig:comp_detectors_training} (2nd column). Computing the shortest distance of a keypoint $z_k=(x_k,y_k)$ to this line, will give us a good intuition of whether this keypoint lies on the object or not. We perform this evaluation, for training detectors with $K=2,3,4$ number of keypoints. We evaluate the detectors on a test dataset with $250$ datapoints, that was held out during training. The distance is computed in pixel space and we report the mean and the standard deviation of that distance. 

\begin{figure}[h]
    \centering
    \includegraphics[trim=30 0 470 20,clip,width=0.3\columnwidth]{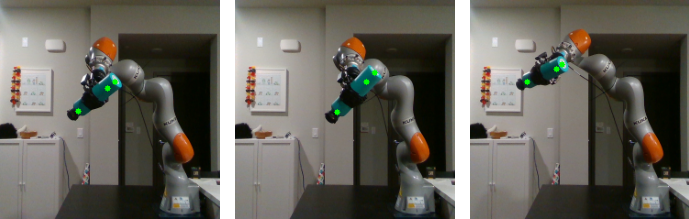}
    \includegraphics[trim=265 0 235 20,clip,width=0.3\columnwidth]{images/experiment_img/hardware_detector_struc_images.png}
    \includegraphics[trim=500 0 0 20,clip,width=0.3\columnwidth]{images/experiment_img/hardware_detector_struc_images.png}\\
    \vspace{1mm}
    \includegraphics[trim=30 0 470 20,clip,width=0.3\columnwidth]{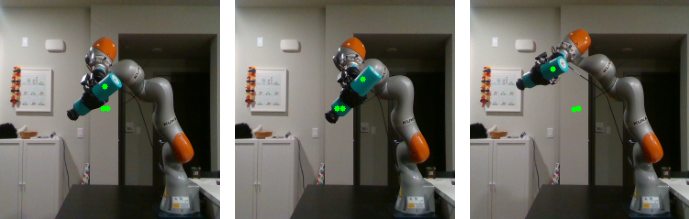}
    \includegraphics[trim=265 0 235 20,clip,width=0.3\columnwidth]{images/experiment_img/hardware_detector_baseline_images.png}
    \includegraphics[trim=500 0 0 20,clip,width=0.3\columnwidth]{images/experiment_img/hardware_detector_baseline_images.png}
    \caption{Qualitative performance of multimodal (top) and baseline (bottom) keypoint detector on the hardware dataset. The multimodal keypoint detector, always detect keypoints on the object, this is not the case for the baseline detector.}
    \label{fig:qualitative_training}
\end{figure}

In Figure~\ref{fig:comp_detectors_training}, we show the averaged distance of each trained keypoint to the line, with error bars, after training. Results for the simulation experiments (top row) and hardware experiments (bottom row) are shown. The bar plots illustrate the distance of the detected keypoints to the center line. We see that the multimodel keypoint detector, that was trained with proprioceptive information, outperforms the baseline detector. The low distance to the center line suggests that the multimodal detector places the keypoints on the object. This is confirmed by the qualitative analysis in Fig. \ref{fig:qualitative_training} where three examples are shown. The multimodal  detector detects keypoints that are on the object (top row). This is not true for the baseline detector, which often fails at detecting keypoints that lie on the object (bottom row).

\subsection{How much proprioception is needed?}
In this experiment, we want to evaluate how combining multi-modal observations, that contain both visual and proprioceptive measurements, with visual only observations affects results. This experiment will give us insights into whether a robot could combine visual observations obtained in a passive way (watching humans in action or videos) with observations gained in an active way (by moving the object itself) to learn keypoint detectors. To perform this experiment, we pretend that for a fraction of observations, the proprioceptive state measurement was not observed, and evaluate how stable keypoint training is as a function of that fraction. 
\begin{figure}
    \centering
    \includegraphics[width=0.4\textwidth]{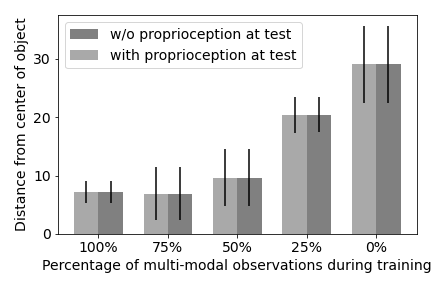}
    \caption{Share of multimodel data (including proprioceptive information) in the dataset used to train the keypoint detector, we compare the performance of the keypoint detectors during test time when including or excluding proprioceptive information. We see that using multimodal data during training shows a significant performance improvement, however during test time the proprioceptive information has no effect on on-object keypoint detection performance.}
    \label{fig:pose_share_barplot}
\end{figure}
As can be seen in Fig. \ref{fig:pose_share_barplot} while we can achieve already good performance when only including $50\%$ of the proprioceptive information during training, the consistency of on-object keypoint detection significantly improves when we include all the proprioceptive information, as can be seen from the smaller standard deviation in the plot. In the plot we can also see that including or not including proprioceptive information during test time has no effect on the performance of detecting keypoints on the object. During test time the keypoint detector can be used only from visual information. This means that our multimodal keypoint detector learned to incorporate the proprioceptive information and has inherently access to it when predicting, even when the proprioceptive information is not available. 
\subsection{Virtual link/joint regression}
In this section we present the experiments showing how we can learn an extended body schema from vision. The purpose of these experiments is to show that we can use our learned visual latent space to regress the translation parameters of some virtual joints that extend the kinematic chain of the robot to also include an object in the manipulator's hand. For this to be successful it is a prerequisite that the visual keypoints are detected on the object, since otherwise the extended kinematic chain would not represent the object in the robot's hand. As we already evaluated in the previous section only our multimodal keypoint detector successfully detects keypoints on the object, therefore we are using our multimodal keypoint detector for these experiments.

We perform two set of experiments for the regression of virtual joint parameters: 1) to evaluate our virtual joint estimation procedure quantitatively, we test whether we can identify ground truth parameters $\Bar{\phi}$, if we knew them; 2) We evaluate regressing virtual links from visual keypoints detected on 4 set of grasps.

\begin{figure}[h]
    \centering
    \includegraphics[width=0.3\textwidth]{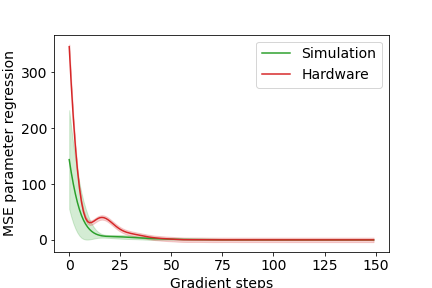}
    \caption{\small Regression of kinematic parameters, MSE to ground truth virtual joint values over gradient steps. Our method is able to regress the virtual joint parameters successfully, on simulation and on hardware. Simulation results are averaged over five seeds and three object with manually chosen ground truth parameters for each object. Hardware results are averaged over five seeds.}
    \label{fig:bar_plot_kin_performace}
\end{figure}

To collect data for the first experiment, we pick a set of three ground truth virtual joint parameters $\Bar{\phi}$ such that their projections $h^k_{\Bar{\phi}}$ lie on the object. We collect a dataset $\mathcal{D} = \{ (\theta_t, h^k_{\Bar{\phi}}(\theta_t)\}_{t=1}^{15}$ for 15 random joint configurations $\theta$. Then, we pretend to not know $\Bar{\phi}$, initialize $\phi$ to be zero, and aim to estimate $\phi$ from the observations in $\mathcal{D}$ via gradient descent, as described in Section~\ref{sec:estimating_joints}. In Fig.\ref{fig:bar_plot_kin_performace}, we show the MSE error between ground truth and estimated parameters $\|\| \phi - \Bar{\phi}\|\|^2$ as a function of number of gradient steps, both from simulation and hardware data. We observe that after approximately $50$ gradient steps $\phi$ converges towards their ground truth values, showing that the regression of virtual joint parameters from projected image and depth values is successful.

\begin{figure}[h]
    \centering
    \includegraphics[trim=50 0 470 20,clip,width=0.3\columnwidth]{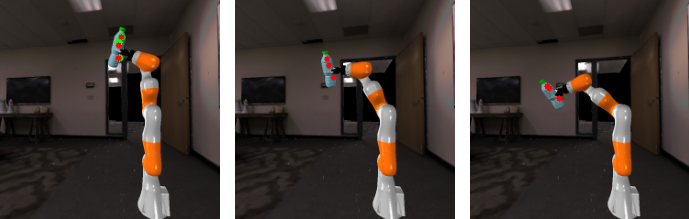}
    \includegraphics[trim=285 0 235 20,clip,width=0.3\columnwidth]{images/experiment_img/kin_regression_images.png}
    \includegraphics[trim=520 0 0 20,clip,width=0.3\columnwidth]{images/experiment_img/kin_regression_images.png}\\
    \vspace{1mm}
    \includegraphics[trim=30 0 470 20,clip,width=0.3\columnwidth]{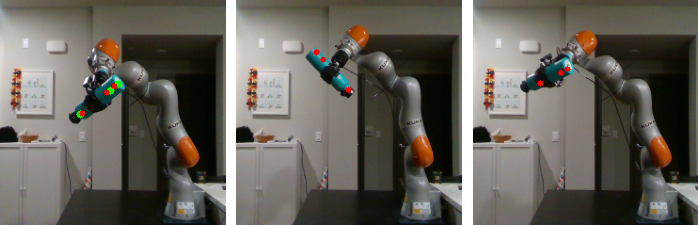}
    \includegraphics[trim=265 0 235 20,clip,width=0.3\columnwidth]{images/experiment_img/hardware_kin_regression_images.png}
    \includegraphics[trim=500 0 0 20,clip,width=0.3\columnwidth]{images/experiment_img/hardware_kin_regression_images.png}
    \caption{\small Learned kinematic extension for three poses for simulation (top) and hardware (bottom). The first images shows, in green, the keypoints detected by the keypoint detector. The red dots are the projections in image given the learned virtual joints. Our method is able to successfully learn parameters $\phi$, that generalize across poses.
    }
    \label{fig:qualitative_kin_training}
\end{figure}

For our second set of experiments, we use the trained keypoint detectors to create a dataset $\mathcal{D} = \{(\theta_t,z_t)\}_{t=1}^{15}$, where $z_t$ are the keypoints predicted by the detector, and learn virtual joints from these visual keypoint observations. The first image of Fig. \ref{fig:qualitative_kin_training} shows the visual keypoints predicted by the multimodal detector (in green), together with the projected virtual links $h^k_\phi$ (in red) after learning $\phi$. The following two images show how the projection of the kinematic chain remains consistent, also when the pose $\theta$ of the manipulator changes. Results are presented for simulation (top) and on hardware (bottom) experiments. The results show that we can successfully learn $\phi$ from visual features. 

\begin{figure}
    \centering
    \includegraphics[trim=55 90 515 20,clip,width=0.3\columnwidth]{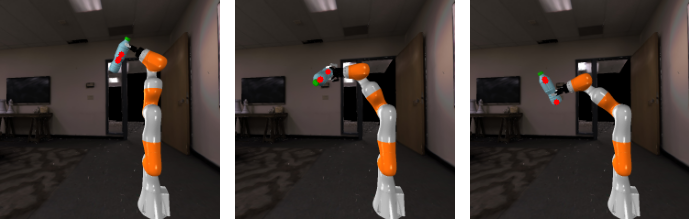}
    \includegraphics[trim=290 90 280 20,clip,width=0.3\columnwidth]{images/experiment_img/kin_regression_images_regrasp.png}
    \includegraphics[trim=525 90 45 20,clip,width=0.3\columnwidth]{images/experiment_img/kin_regression_images_regrasp.png}
    \caption{Re-grasping for three different grasps. After each new grasp we regress the kinematic parameters successfully.}
    \label{fig:kin_regrasp}
\end{figure}

Finally, we also evaluate our kinematic parameter regression when the robot re-grasps the object. Re-grasping changes the kinematic chain, by for example shifting it. In Fig. \ref{fig:kin_regrasp}, we show how our method can successfully learn kinematic parameters $\phi$ that reflect this change. We show results for three new grasps. In the next section we present results for a downstream placing task, where our learned extended kinematic model is used, with different grasps, to place the object on a table.
\section{Down-stream Task Experiments: Model-based Control for Object Manipulation}
\begin{figure*}
    \centering
    \includegraphics[width=0.3\textwidth]{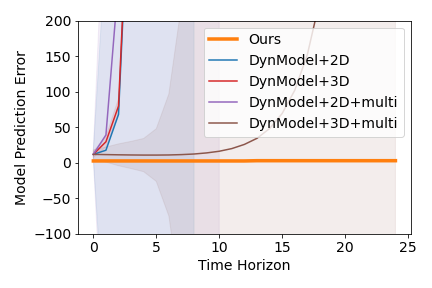}
    \includegraphics[width=0.3\textwidth]{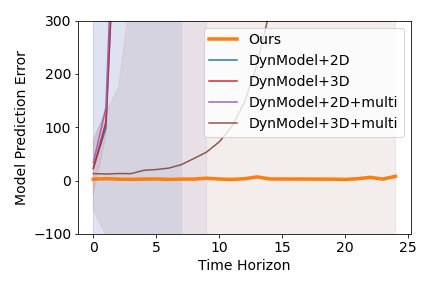}
        \includegraphics[width=0.3\textwidth]{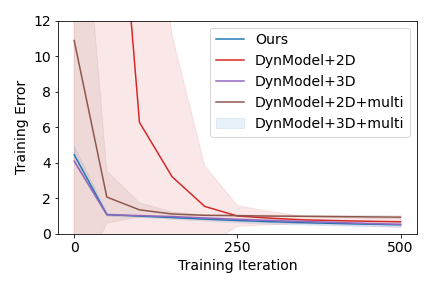}
    \caption{\small On the two plots on the left the long horizon prediction performance of our approach compared with neural networks dynamics models is shown. The plot on the left shows the error when using the action sequence that performs a placing task, the plot in the middle shows the prediction error for a random action sequence.  The prediction error of the neural network increases exponentially over time or is high to begin with, while the error for our method stays small. The plot on the right shows the training error of the neural network models, to show that they were trained until convergence. We report the mean and the standard deviation of the error.}
    \label{fig:long_horizon_pred}
\end{figure*}
\begin{figure}
    \centering
    \includegraphics[trim=50 0 50 100,clip,width=0.24\columnwidth]{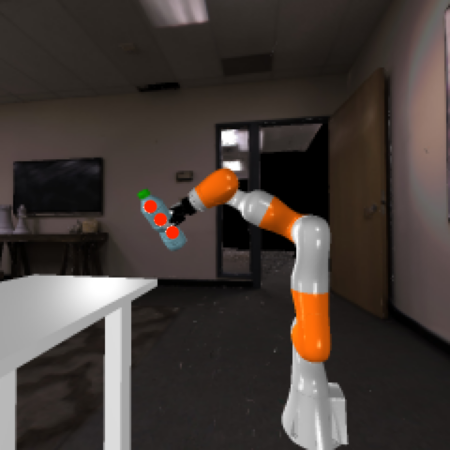}
    \includegraphics[trim=50 0 50 100,clip,width=0.24\columnwidth]{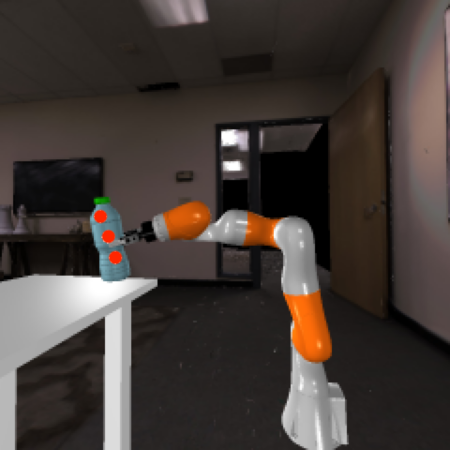}
    \includegraphics[trim=25 0 100 125,clip,width=0.24\columnwidth]{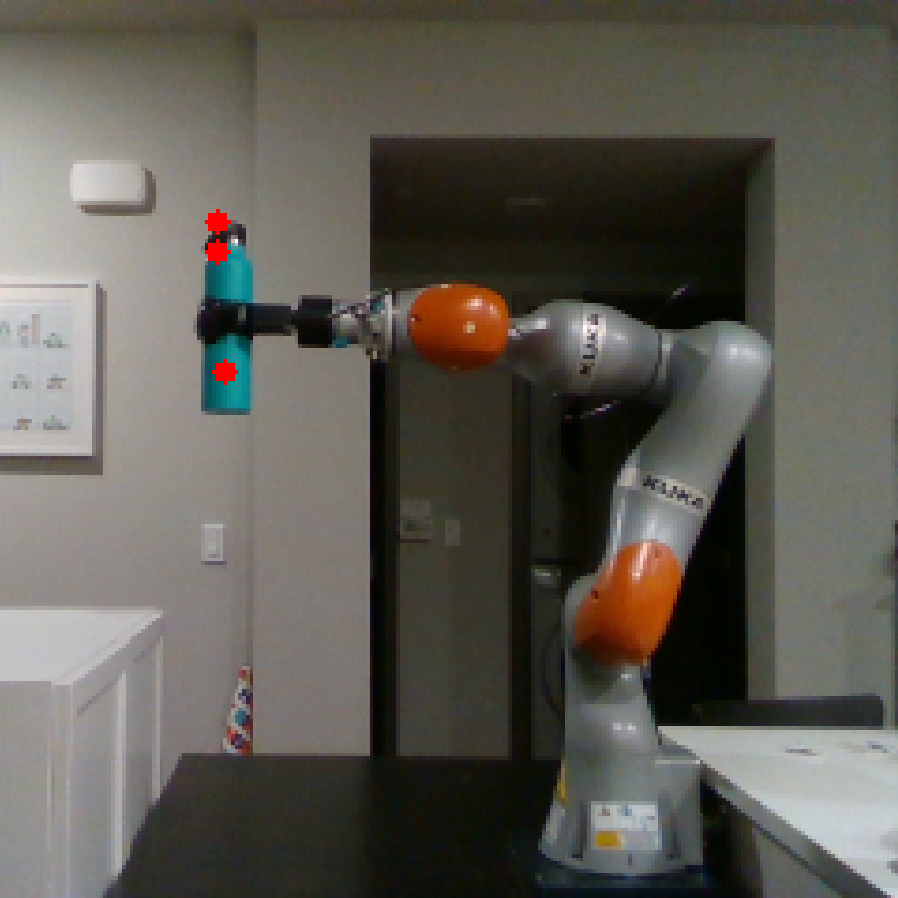}
    \includegraphics[trim=25 0 100 125,clip,width=0.24\columnwidth]{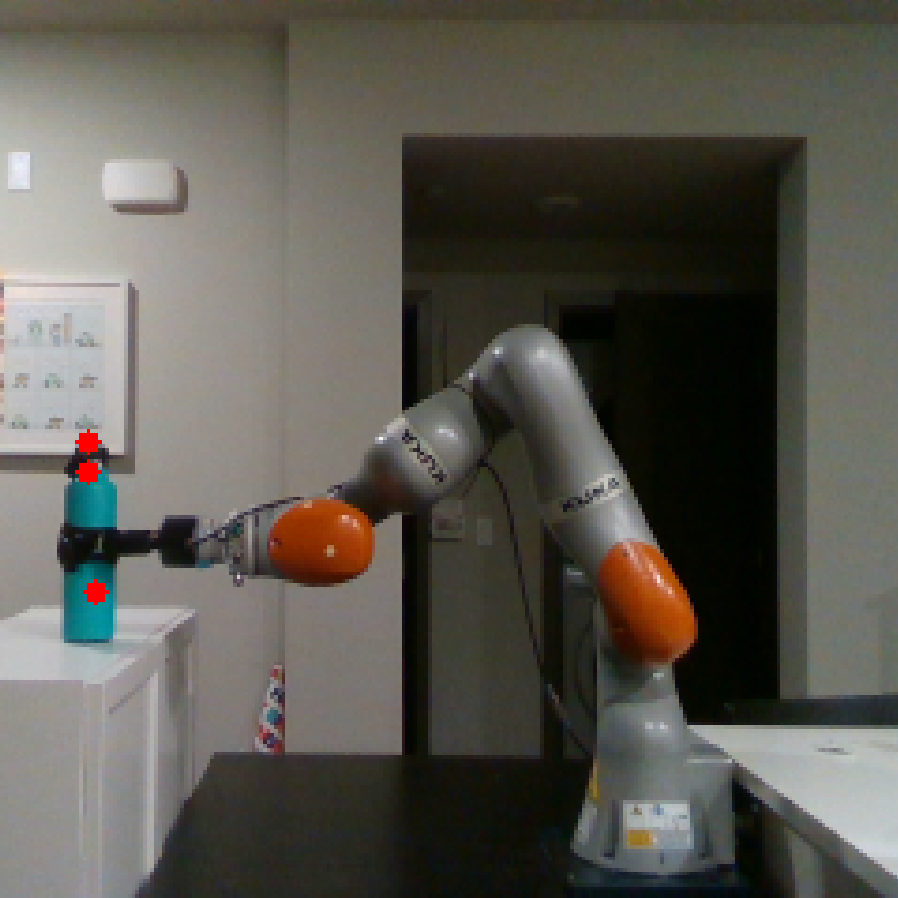}
    \caption{Placing task in simulation and hardware}
    \label{fig:placing_task_hardware}
\vspace{-15pt}
\end{figure}
Finally, we want to use our approach of learning extended body schemas from multimodal data on a object manipulation task, specifically a placing task (see Figure~\ref{fig:placing_task_hardware}). With these experiments we want to show the practical applicability of our approach on a robotic task, and present results in simulation as well as on hardware. We compare our approach to various baselines. 

During the placing task, the robot is required to manipulate the object in the gripper such that it successfully places it on a table. The task is defined in the visual domain, with a desired goal image and associated keypoint positions in pixel. The optimization also takes place in keypoint-space.

Performing a motor control task, such as a placing task, in a model based fashion, requires optimizing controls using a model of the robot and the task. In the case of the placing task the model needs to capture information about the arm movement, when an action is applied, and the corresponding movement of the object in the manipulator's hand. The actions are optimized for a time-horizon of $T=10$ time steps, with the gradient based approach introduced in section \ref{sec:gradient_based_control}. We define a cost function, which penalizes the distance between predicted and goal keypoint locations. We compare our learned extended kinematic chain to various baselines that have previously been presented in \cite{neha,manuelli2020keypointfuture}. The baselines also use visual keypoint detectors for the latent visual representation $z$ but, instead of learning an extended kinematic chain, learn a neural network black-box dynamics model $g_\beta$, $s_{t+1} = g_\beta(s_t,u_t)$ that maps the current state $s_t$ and action $u_t$ to the next state $s_{t+1}$. As previously $s_{t} = [\theta_t,z_t]$. 

We show results for experiments in simulation and on hardware. Our approach learns an extended kinematic chain from the multimodal visual keypoints detector, that uses proprioceptive information during training (see section \ref{sec:approach}). The virtual joints, that represent the object, are regressed from a few measurements, as we described previously. Specifically we used three keypoints and thus, regress the parameters of three virtual joints.  We compare our approach to the following four baselines that learn a neural network dynamics model in the keypoint latent space:
\paragraph{DynModel in 2D Keypoint space (DynModel+2D)} We train a keypoint detector as described in \cite{minderer2019unsupervised, neha} without depth information. This keypoint detector does not use any proprioceptive information during training. We use the keypoint detector to train a dynamics model $s_{t+1} = g_\beta(s_t,u_t)$ from a collected dataset, this reimplements the approach presented in \cite{neha}. 
\paragraph{DynModel in 3D Keypoint space (DynModel+3D)} This baseline extends the baseline presented in a) by including also depth information during keypoint detector and dynamics model training. The latent space now is three dimensional as it includes $(x, y, d)$ - pixel locations and depth information. With this baseline we want to test whether adding depth information improves task performance.
\paragraph{DynModel in 2D-multimodal keypoint space (DynModel+2D+multi)} We train a structured keypoint  detector as described in section \ref{sec:structured_keypoint_training}, the structured keypoint detector is trained using proprioceptive information. After keypoint detector t raining we train a black box neural network dynamics model $s_{t+1} = g_\beta(s_t,u_t)$, instead of regressing virtual joint parameter. This baseline will allow us to see whether using the structured keypoint detector also improves performance on the manipulation task, even if the visual dynamics model is learned with a neural network. 
\paragraph{DynModel in 3D-multimodal keypoint space DynModel+3D+multi)} same as above the baseline in c) but here again we include depth information for keypoint detector training and neural network dynamics model training. As in the baseline before we want to test whether using proprioceptive information during keypoint detector training improves performance on the task, but we also want to test whether adding depth information, and thus a third dimension to our visual latent space, improves performance. 

For all our baselines we train a visual dynamics model in the keypoint space represented by a neural network. We use dataset $\mathcal{D}=\{(\theta_t,u_t,z_t))\}_{t=0}^{2000}$ collected on sine motions of the robot, where $z$ are the keypoints predicted by the detectors, to train a feedforward neural network. All the dynamics models are trained to convergences on the training data, and achieve a normalized mean squared error below $0.1$ on the test data. Training a visual dynamics model in the latent space has been a popular choice in the literature so far, \cite{neha,manuelli2020keypointfuture}. It has the advantage that neural networks have a lot of flexibility to represent the problems at hand, however a downside of neural network based approaches is the unreliable extrapolation behaviour for out of distribution data and long horizon prediction performance, that degenerates fast and is detrimental for model-based control.
\begin{table}[H]
\centering
		\begin{footnotesize}
			\begin{tabular}[b]{p{0.5 cm} c c c c c c c}
				\toprule
				\bf Method & \bf Task 1 & \bf Task 2 & \bf Task 3\\ 
				& Mean (Std) & Mean (Std) & Mean (Std) \\
				\midrule
				Ours & \bf 4.85(2.91) & \bf 2.04(1.26) & \bf 2.06(1.18) \\
				a & 155.62(343.84) &60.78(73.67) & 70.00(67.73)\\
				b & 135.16(119.7) &69.60(59.69) & 85.32(83.27) \\
				c & 103.35(85.07) &29.37(48.09) & 70.38(88.42) \\
				d & 120.21(107.39) &40.14(42.95) &96.58(258.14) \\
				\bottomrule
				Hardware & \ \bf 4.06 (3.83) & \bf 2.12 (0.04) & \bf 5.60 (4.43) \\
				\bottomrule
			\end{tabular}
		\end{footnotesize}
		\vspace{0.1cm}
		\label{fig:placing_task}
		\caption{ \small The performance of three placing tasks. We compare our method to the baselines introduced before. We report the final distance of the object in the hand from the desired target position of the object as root mean squared error. We average our results over five seeds and four different re-grasps of the object in simulation. On hardware we average over five seeds. The errors are reported in pixel space.}
		\label{table:placing}
\end{table}
In Fig. \ref{fig:long_horizon_pred} we show the predictive behaviour of our baseline neural network models. In the image on the right, the training error of the models is shown and it is clearly possible to see that the models were trained until convergence. Nevertheless the two plots on the left show the long horizon prediction behaviour of the neural networks compared with out extended kinematic chain. On the x axis the time horizon is shown and on the y axis the prediction error. From the plots it becomes evident that predicting for longer horizons through the neural network dynamics models, which is necessary when optimizing actions for a motor control task like placing, leads to exponential prediction error. The plot on the left shows long horizon prediction for the placing task action sequence, the plot in the middle for a random action sequence. In all cases the predictive performance of the neural networks deteriorates exponentially over time for all baselines which also explains the poor results presented in Table \ref{table:placing} where we report the results for our experiments. Only the neural network model that is trained using our multimodal keypoint detector with depth information (baseline d), can achieve a satisfactory predictive behaviour over 10 time steps, showing that improved on-object keypoint detection can increase also the performance of neural network visual dynamics models. 

In Table \ref{table:placing} the simulation experiments are averaged over five seeds, the three objects introduced earlier and four different grasp positions. The results show the final distance of the object keypoints in the manipulators hand to the desired goal keypoints on the table. We perform a total of three tasks that vary in start-goal configurations. 
All the distances are reported in pixel space, and represents the root mean squared error (RMSE). Between all the baselines, the baselines c) and d) that use our proposed keypoint detector, perform best. This indicates, that the keypoints trained with our detector are better suited for the object manipulation task. Yet overall, compared to our extended kinematic chain, all of them perform fairly poorly in the model-based control task. Our approach outperforms all baseline variations and is the only one to successfully place the object on the table. This gap in performance can be explained by the fact that the extended kinematic is able to generalize throughout the state-space, while the trained dynamics models prediction quality deteriorates quickly outside of the training data distribution. 

It is worthwhile highlighting, that our approach is successful also when used on the real robot, where we achieve an average RMSE of $3.92$ pixels on the placing task. This is remarkable not only because on hardware noise and controller inaccuracies often make the task harder, but also because we run our optimized action sequence in a feedforward fashion, without having to replan at each time step.

\section{Discussion and Future Work}
We present a method for learning extended body schemas from vision. Our method inherently incorporates an external object in the manipulator's hand. To this end we show how merging two sensor modalities, precisely proprioceptive information, in form of joint positions, and vision, enables us to learn better visual latent representation of the object. The latent representation is then used to successfully learn an extension of a robot's kinematics model. As a result the learned extended kinematic chain incorporates the object into the kinematic model of the robot. We show how we can use the learned kinematic model for object manipulation on a placing task, where the task is defined in the visual domain. We also show how we can adapt the kinematic chain for different grasps from only few datapoints. Our experiments, on a 7 DoF iiwa Kuka arm in simulation and on hardware, show the generality of our approach and the good performance to out of distribution tasks and on the real system. In contrast to current state of the art, we leverage structured analytical models, i.e the kinematic model, and combine this structure with a data driven approach to learn visual latent keypoint representations for consistent on-object keypoint placing. When learning the extended kinematic chain, we again combine structure and learning to recover a representation on the robot's arm with the object that transfers easily to new tasks. We believe the combination of data driven learning and structured models are an interesting avenue for further research. A remaining challenge of our approach is how to decide when to add or remove an object from the extended kinematic chain. This depends on the grasp and release behaviour of the desired task. In the future we would also like to explore more complex, contact rich manipulation tasks, that use the learned extended body schema to control the robot during contact rich manipulation. We also believe that combining vision only data with robot experience in form of proprioception is an interesting avenue for future work. This would allow to leverage larger visual dataset, not collected directly on the robot, and combine this visual data with a smaller but informative amount of robot experience. 
% use section* for acknowledgment
% \section*{Acknowledgment}

% The authors would like to thank...

% Can use something like this to put references on a page
% by themselves when using endfloat and the captionsoff option.
\ifCLASSOPTIONcaptionsoff
  \newpage
\fi

% trigger a \newpage just before the given reference
% number - used to balance the columns on the last page
% adjust value as needed - may need to be readjusted if
% the document is modified later
%\IEEEtriggeratref{8}
% The "triggered" command can be changed if desired:
%\IEEEtriggercmd{\enlargethispage{-5in}}

% references section

% can use a bibliography generated by BibTeX as a .bbl file
% BibTeX documentation can be easily obtained at:
% http://mirror.ctan.org/biblio/bibtex/contrib/doc/
% The IEEEtran BibTeX style support page is at:
% http://www.michaelshell.org/tex/ieeetran/bibtex/
%\bibliographystyle{IEEEtran}
% argument is your BibTeX string definitions and bibliography database(s)
%\bibliography{IEEEabrv,../bib/paper}
%
% <OR> manually copy in the resultant .bbl file
% set second argument of \begin to the number of references
% (used to reserve space for the reference number labels box)

\bibliographystyle{IEEEtran}
\bibliography{root}

% biography section
% 
% If you have an EPS/PDF photo (graphicx package needed) extra braces are
% needed around the contents of the optional argument to biography to prevent
% the LaTeX parser from getting confused when it sees the complicated
% \includegraphics command within an optional argument. (You could create
% your own custom macro containing the \includegraphics command to make things
% simpler here.)
% ve a space for a photo:

% \begin{IEEEbiography}{Michael Shell}
% Biography text here.
% \end{IEEEbiography}

% % if you will not have a photo at all:
% \begin{IEEEbiographynophoto}{John Doe}
% Biography text here.
% \end{IEEEbiographynophoto}

% insert where needed to balance the two columns on the last page with
% biographies
\begin{IEEEbiography}[{\includegraphics[width=1in,height=1in,clip,keepaspectratio]{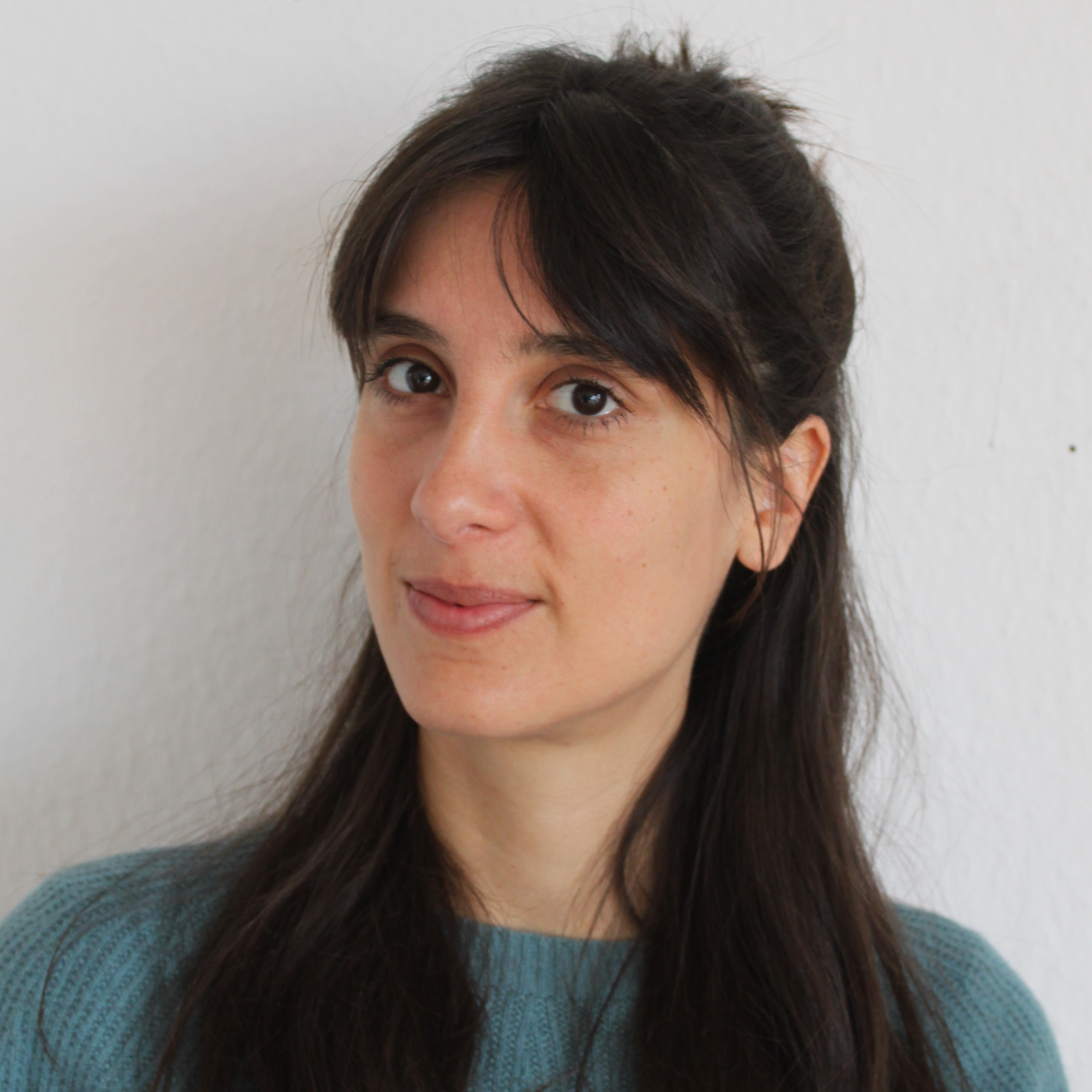}}]{Sarah Bechtle}
is a last year PhD candidate at the Max Planck Institute for intelligent Systems in Tuebingen. She is also affiliated with the Machines in Motion Lab at NYU. Previously she received a Master’s degree in computational neuroscience from the Bernstein Center in Berlin and a Bachelor’s degree in computer science from the University of Munich. 
Her research interests are at the intersection between machine learning and robotics. Specifically she is interested in model based learning within the action-perception-learning loop as well as in meta and lifelong learning. 
\end{IEEEbiography}

\begin{IEEEbiography}[{\includegraphics[width=1in,height=1in,clip,keepaspectratio]{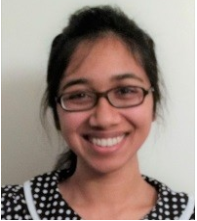}}]{Neha Das}
is a first year PhD candidate at the Chair of Information Oriented Control at the Technical University of Munich.  Previously she received a Master’s degree in Informatics from Technical University of Munich and a Bachelor’s degree in Software Engineering from Delhi Technical University (formerly Delhi College of Engineering) in India. 
Her research interests are at the intersection between machine learning and control. She is interested in modelling dynamical systems and prediction, as well as model predictive control and meta-learning methods in those contexts. 
\end{IEEEbiography}

\begin{IEEEbiography}[{\includegraphics[width=1in,height=1.25in,clip,keepaspectratio]{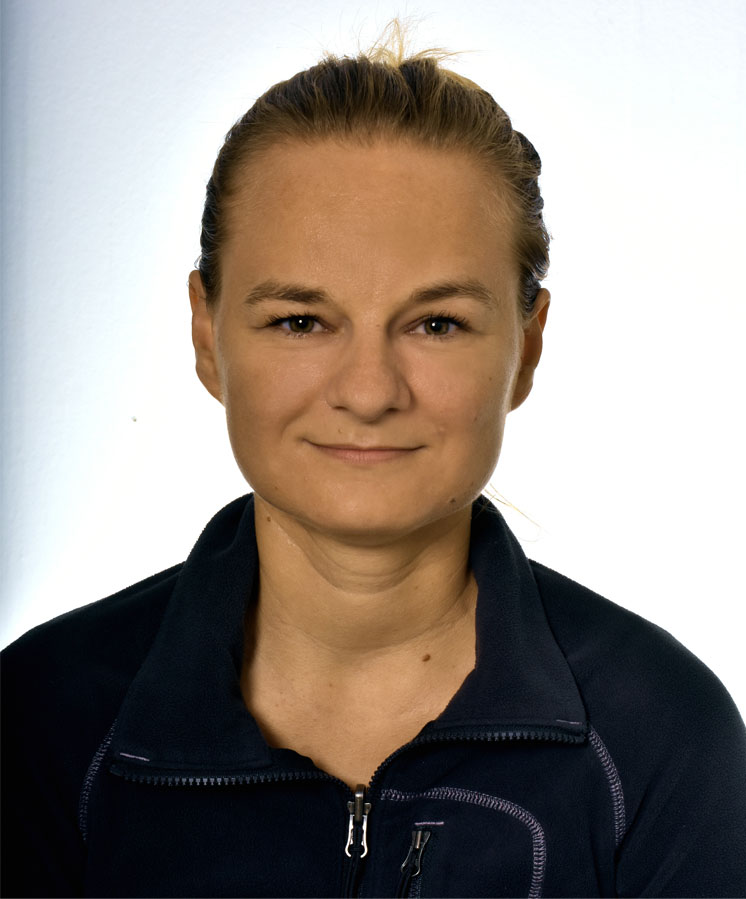}}]{Franziska Meier}
Franziska Meier is a research scientist at FAIR (Facebook AI Research). Previously she was a research scientist at the Max-Planck Institute for Intelligent Systems and a postdoctoral researcher with Dieter Fox at the University of Washington, Seattle. She received her PhD from the University of Southern California, where she defended her thesis on “Probabilistic Machine Learning for Robotics” in 2016, under the supervision of Prof. Stefan Schaal. Prior to her PhD studies, she received her Diploma in Computer Science from the Technical University of Munich. Her research focuses on machine learning for robotics, with a special emphasis on lifelong learning for robotics.
\end{IEEEbiography}

\end{document}